\documentclass[sigconf]{acmart}
\usepackage{subfigure}
\usepackage{CJKutf8}
\newcommand{\multirows}[1]{\begin{tabular}{@{}c@{}}#1\end{tabular}}

\AtBeginDocument{%
  \providecommand\BibTeX{{%
    \normalfont B\kern-0.5em{\scshape i\kern-0.25em b}\kern-0.8em\TeX}}}

\setcopyright{rightsretained}
\begin{document}
\fancyhead{}

\title{DSGPT: Domain-Specific Generative Pre-Training of Transformers for Text Generation in E-commerce Title and Review Summarization}

\author{
Xueying Zhang, 
Yunjiang Jiang, 
Yue Shang, 
Zhaomeng Cheng, 
Chi Zhang,
Xiaochuan Fan, \\
Yun Xiao,
Bo Long
}


\affiliation{%
  \institution{JD.com Silicon Valley Research Center}
  \city{Mountain View}
  \state{CA}
  \country{USA}
}


\begin{abstract}
We propose a novel domain-specific generative pre-training (DSGPT) method for text generation and apply it to the product title and review summarization problems on E-commerce mobile display. 
First, we adopt a decoder-only transformer architecture, which fits well for fine-tuning tasks by combining input and output all together.
Second, we demonstrate utilizing only small amount of pre-training data in related domains is powerful. 
Pre-training a language model from a general corpus such as Wikipedia or the Common Crawl requires tremendous time and resource commitment, and can be wasteful if the downstream tasks are limited in variety.
Our DSGPT is pre-trained on a limited dataset, the Chinese short text summarization dataset (LCSTS).
Third, our model does not require product-related human-labeled data. 
For title summarization task, the state of art explicitly uses additional background knowledge in training and predicting stages. 
In contrast, our model implicitly captures this knowledge and achieves significant improvement over other methods, after fine-tuning on the public Taobao.com dataset.
For review summarization task, we utilize JD.com in-house dataset, and observe similar improvement over standard machine translation methods which lack the flexibility of fine-tuning.
Our proposed work can be simply extended to other domains for a wide range of text generation tasks.
\end{abstract}



\copyrightyear{2021} 
\acmYear{2021} 
\acmConference[SIGIR '21]{Proceedings of the 44th International ACM SIGIR Conference on Research and Development in Information Retrieval}{July 11--15, 2021}{Virtual Event, Canada}
\acmBooktitle{Proceedings of the 44th International ACM SIGIR Conference on Research and Development in Information Retrieval (SIGIR '21), July 11--15, 2021, Virtual Event, Canada}\acmDOI{10.1145/3404835.3463037}
\acmISBN{978-1-4503-8037-9/21/07}

\maketitle

\section{Introduction}
In recent years, online transactions are made on mobile phones more and more frequently than on PCs.
Improving user experience in mobile Apps or mini-programs within mobile Apps has become very important for E-commerce companies. 
Text summarization plays a significant role due to the limitation of mobile screen size. 
Here we focus on two key problems: title shortening and review highlighting. 
In some cases, the product titles on the E-commerce platform provided by the sellers are redundant, difficult to read, or exceed the limitation of screen size. As shown in Figure 1, the product title has more than 70 Chinese words and lots of redundant information, which makes it hard to be displayed on the search page to reflect all useful information. This may hurt user's search experience seriously. 
Besides title and product information, product reviews provide users additional information from customers' perspective. To better assist users to find the right product while searching, displaying key information of product reviews has become a trending feature. As shown in Figure 2, the product reviews are often too long to be displayed on the search page. 
To provide customers better search experiences in online shopping, generation of short and informative summarization for title and review has become an important and essential research problem in E-commerce. 

\begin{figure}[htbp]
    \centering
    \subfigure{
        \includegraphics[width=1.2in]{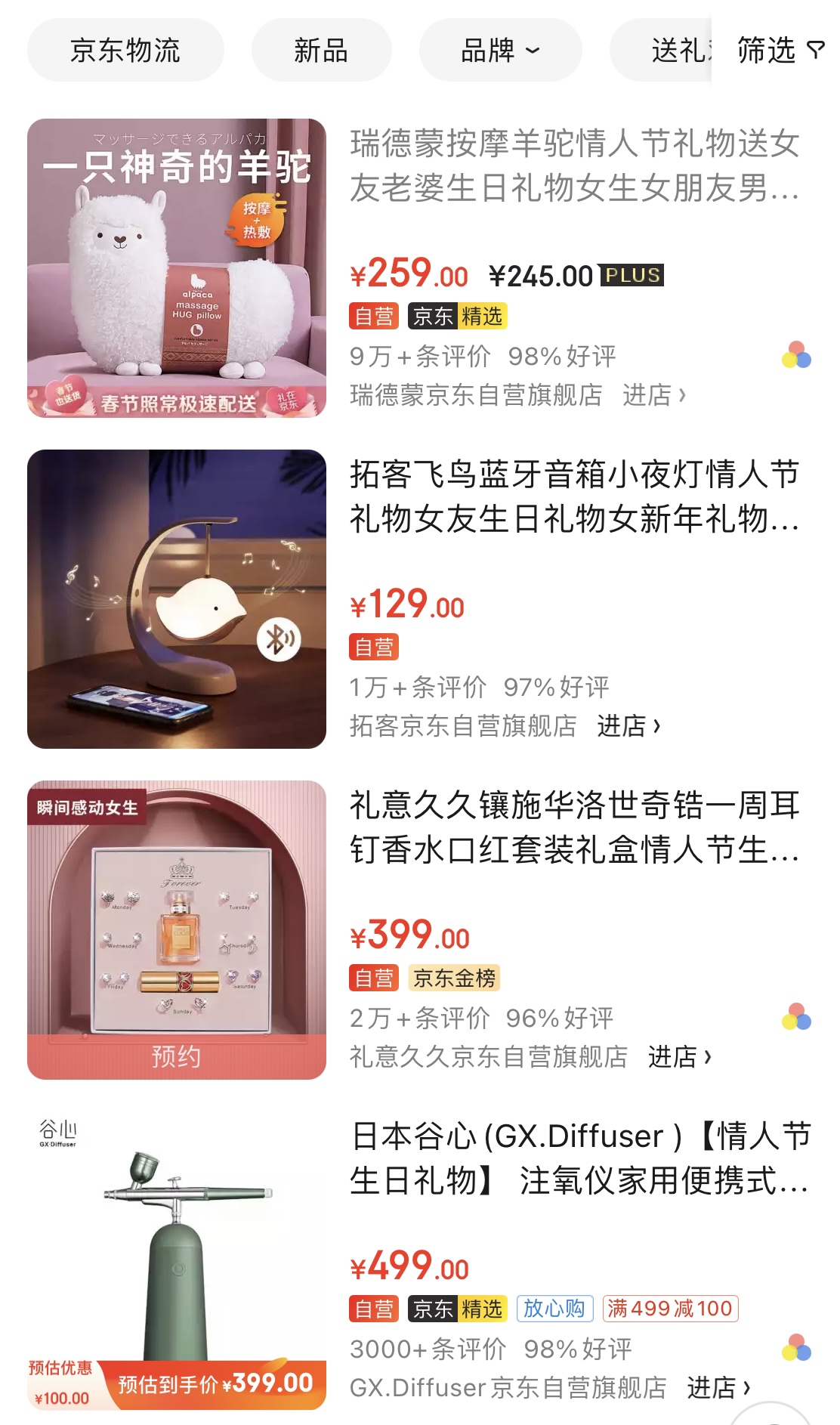}
    }
    \subfigure{
        \includegraphics[width=1.2in]{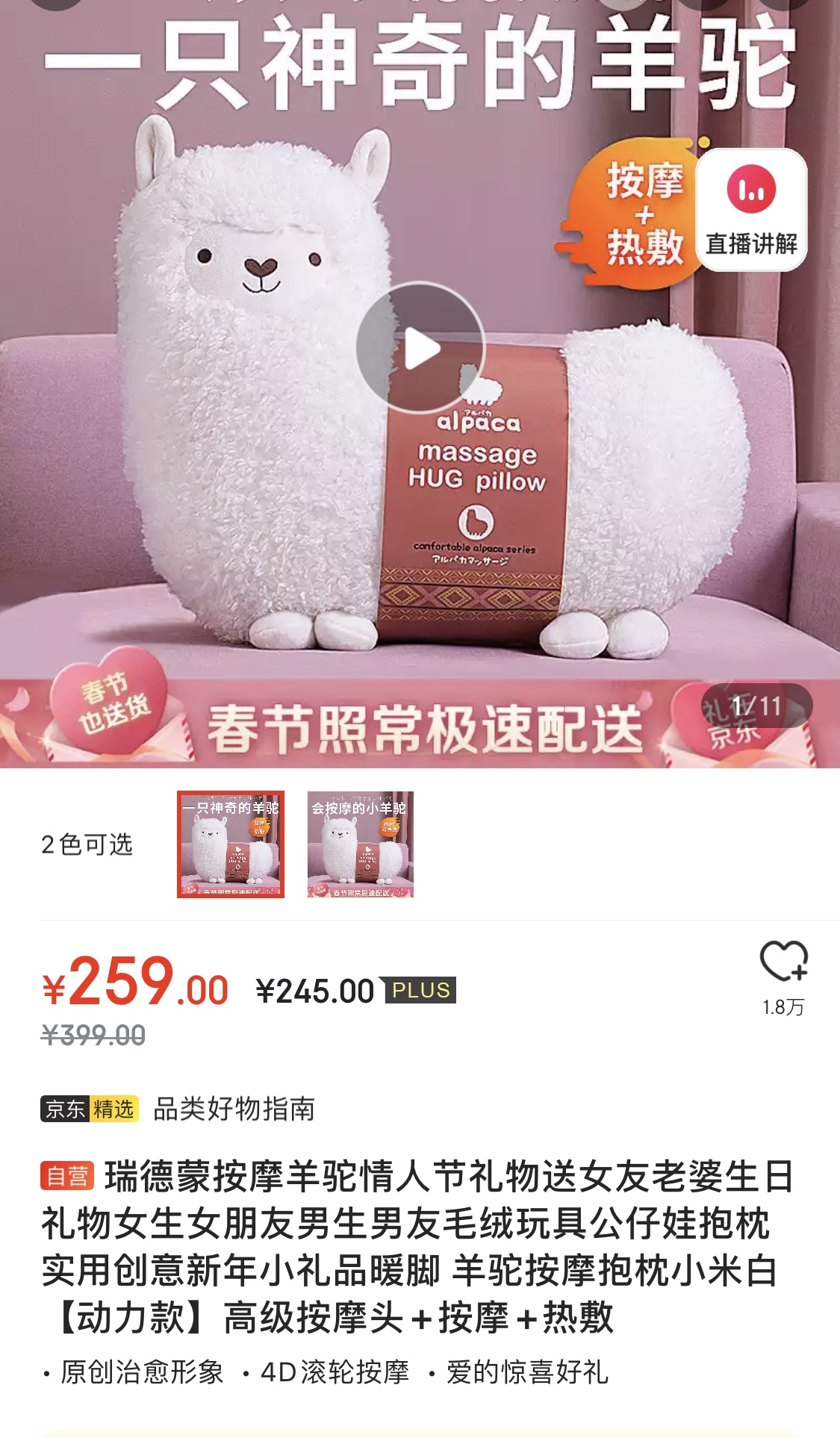}
    }
    \caption{The product titles are truncated in order to fit within the e-commerce search mobile APP.}
\end{figure}

Since the introduction of BERT\cite{devlin-etal-2019-bert} in 2018, pre-training and fine-tuning techniques have been widely used in natural language processing tasks.
In particular, this 2-stage learning process has been applied to many different kinds of downstream tasks such as classification, named entity recognition, question-answering, etc. Pre-trained models like BERT are open-sourced in multiple languages and different sizes for ease of use in fine-tuning tasks. However, due to the bidirectional training methodology in pre-training process, these models cannot be utilized for fine-tuning in text generation tasks. Open-sourced pre-trained models which can be used for text generation like GPT-2/3\cite{radford2019language}\cite{brown2020language}, CTRL\cite{keskar2019ctrl} are too large for real-world applications and in English only. 

\begin{figure}[htbp]
    \centering
    \subfigure{
        \includegraphics[width=1.2in]{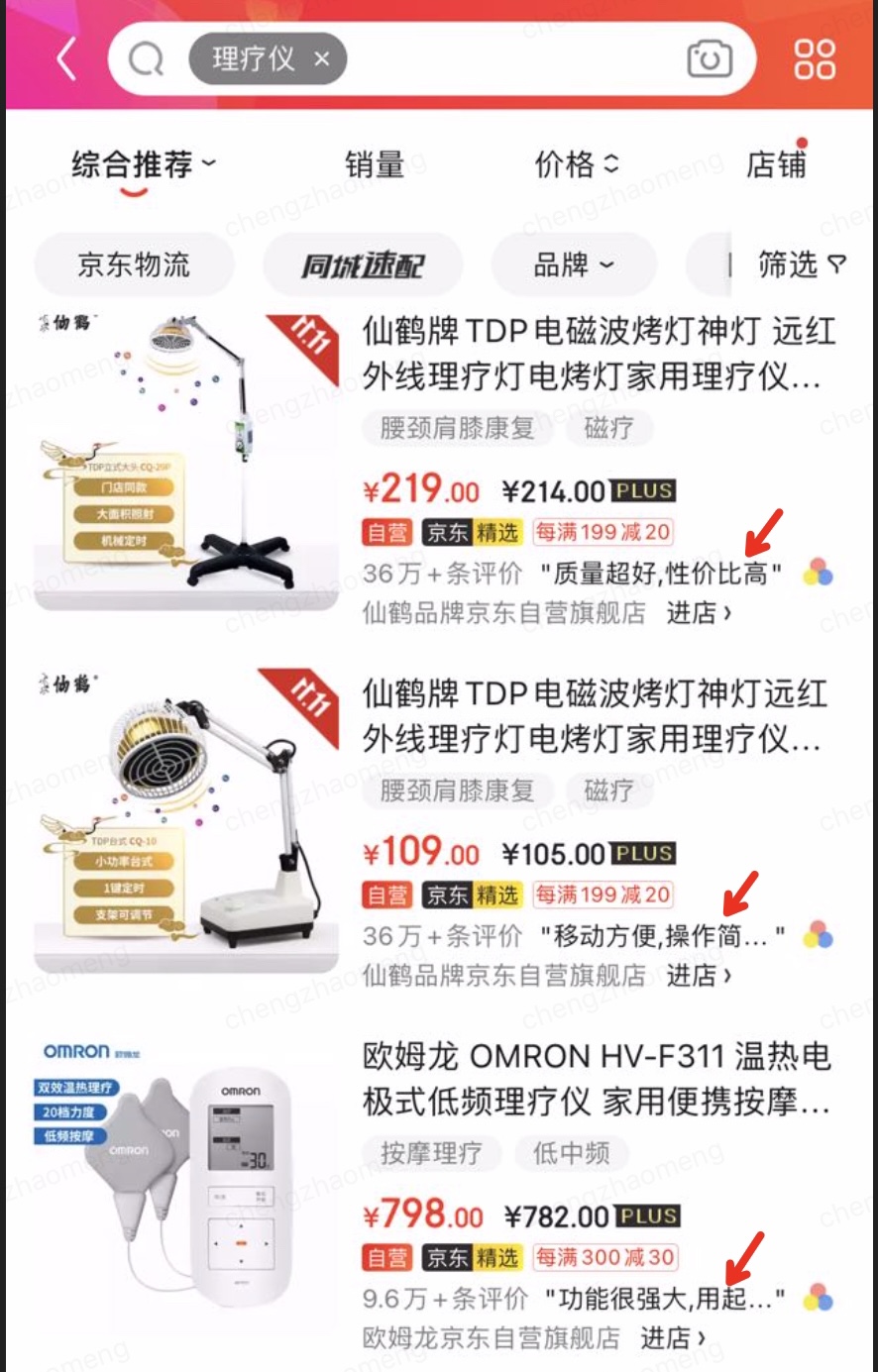}
    }
    \subfigure{
        \includegraphics[width=1.2in]{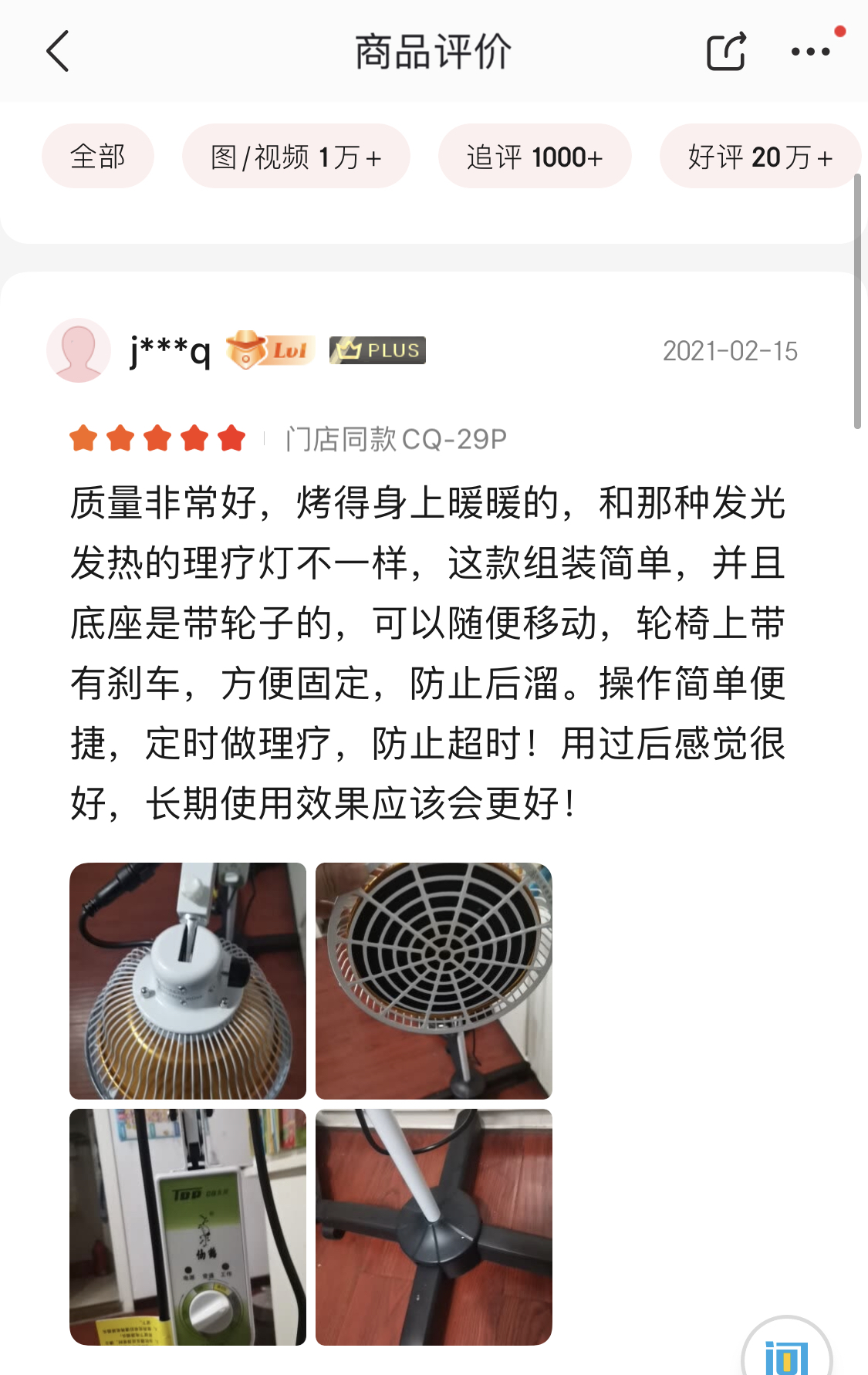}
    }
    \caption{Review highlight example in the E-commerce search.}
\end{figure}  

To overcome these challenges, we propose a domain-specific generative pre-training method, which expands the widely applied idea of pre-training and fine-tuning to the field of text generation while utilizing only small amount of data in related domains. We utilize decoder-only transformer structure and demonstrate that in some situations, this structure can achieve better performance than encoder-decoder transformer model due to the following reasons:

\begin{itemize}
\item The decoder-only structure makes it easy to feed input and output data together. There is no need to change the model structure when applying the model to new data, making it a great choice for pre-training and fine-tuning tasks.
\item The decoder-only structure is easy to scale up, and the performance improves significantly when model becomes larger.
\item Different tasks can be distinguished by adding special tokens to training data.
\item Earlier studies show that pre-trained models need super large amount of data and large model size to obtain significant improvements. We demonstrate that only relevant data is needed in pre-training task, and small amount of data and regular-size model also improve the performance a lot. 
\item Generalized background knowledge can be learned implicitly in pre-training process, fine-tuned models can achieve good performance without utilizing additional human-labeled explicit background knowledge in downstream tasks. 
\end{itemize}

\section{Related Work}

Automatic text summarization methods can be categorized into two paradigms: the extractive summarization and the abstractive summarization. In extractive summarization, important words or sentences are extracted from the original document and then arranged to form a summary. The important units of text can be detected through classification methods\cite{kupiec1995trainable}, selection methods\cite{nallapati2017summarunner} and ranking methods\cite{gong2001generic}. 

In abstractive summarization, the source document is rewritten into a shortened version which preserves the core meaning of the original text\cite{rush2015neural}. In early studies, abstractive summarization was realized by statistical machine translation techniques\cite{michele2000headline}\cite{kevin2000statistics}. With the development of deep learning, NN-based encoder-decoder frameworks such as attentional Seq2Seq models\cite{rush2015neural} and Transformers\cite{vaswani2017attention} have pushed abstractive summarization to new states of the art.  

The OpenAI GPT-2 and GPT-3 \cite{radford2019language}\cite{brown2020language} are large-scale language models pre-trained on large datasets that can achieve state of the art results via fine-tuning, zero-shot or few-shot. Salesforce research team has proposed CTRL\cite{keskar2019ctrl}, which utilizes control codes to realize text generation with different styles and task-specific behaviors. Microsoft has proposed a task-oriented pre-trained model integrated in the dialog system DialoGPT\cite{zhang2020dialogpt}.  The open-domain and task-oriented pre-trained models discussed above are all obtained with large amounts of data and compute resources. In this paper, we demonstrate that with small amounts of data and small models we can also get benefits from pre-training.

\begin{figure}[htbp]
    \centering
    \includegraphics[width=2in]{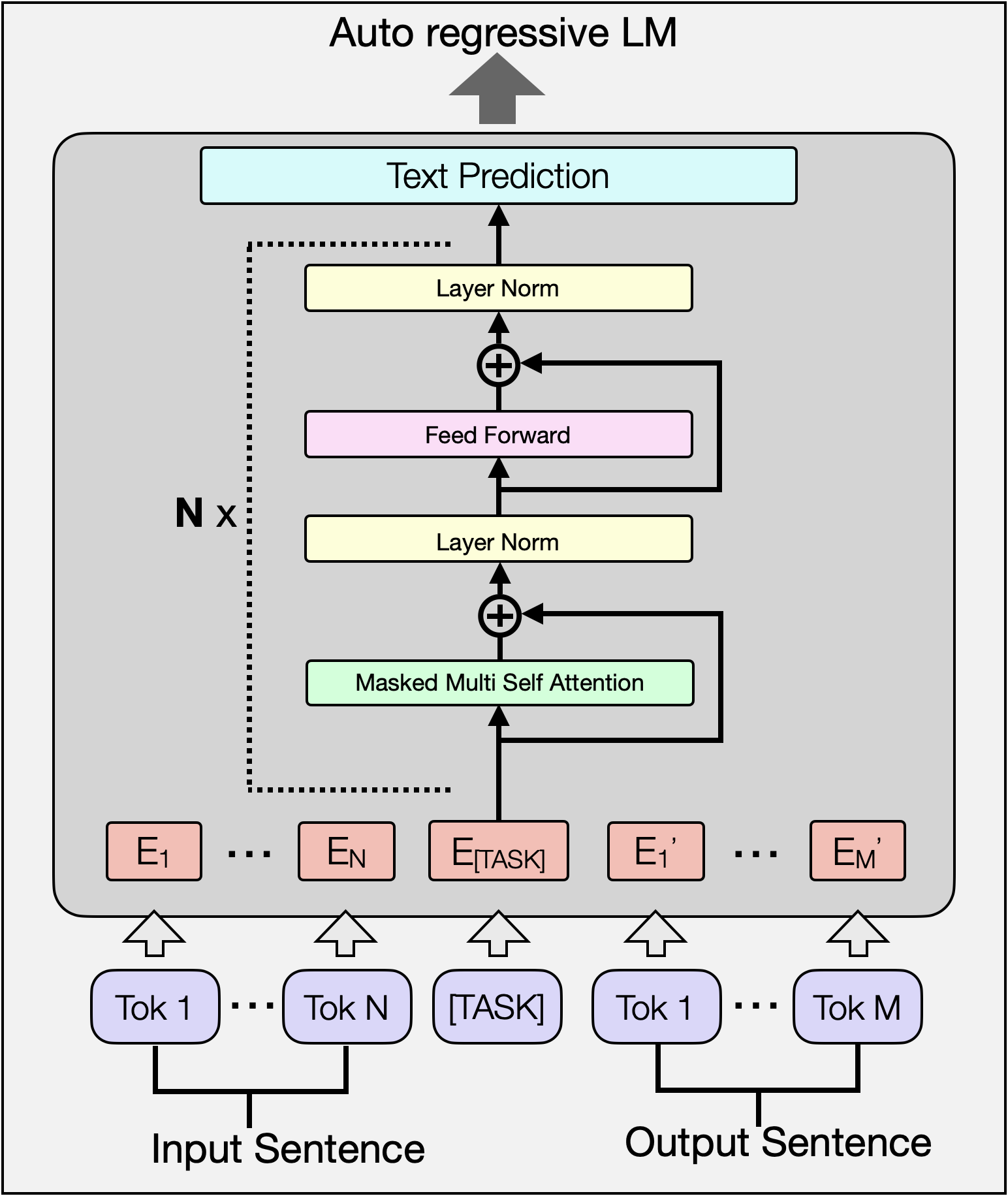}
    \caption{Model Structure}
\end{figure}  

\section{Approach}

\subsection{Method}

Recall that the core of text generation is language modeling.  
Given a set of examples each consisting of a sequence of symbols with variable lengths $(s_1; s_2; ...; s_n)$, language modeling is to estimate the probability distribution of the corpus. 
The joint probabilities over symbols are factorized as the product of conditional probabilities\cite{bengio2003neural}:

\begin{equation}
P(x) = \prod_{i = 1}^n P(s_n|s_1, ..., s_{n-1}). 
\end{equation}

Language can specify inputs and outputs as a sequence of symbols in a flexible way. In this way, one model with the same structure can be reused for different tasks by only changing the input data which contains information of inputs, outputs and tasks all together.

\subsection{Domain-Specific Generative Pre-training}

Unlike machine translation problem which includes encoder and decoder, our model has decoder only. 
As shown in Figure 3, the input, output and task are fed into the decoder transformer as the new input of the auto-regressive language model.
This simplification makes it easy to handle multiple tasks in one model. To enable further use for multiple tasks within one model structure, we control the structure of the input data. In summarization problem, we separate input text and output summarization with a special token, and feed the merged sequence of words into language model directly. The target of the pre-training process is to minimize the perplexity of language model on domain-data.

\subsection{Task-Specific Fine-Tuning}

Once the pre-trained model in specific domain is obtained, (our domain is text summarization in Chinese), similar tasks with less data can be fine-tuned from it. The fine-tuning process is similar with the pre-training process, except that the weights are initialized from the pre-trained model. The fine-tuned model inherits knowledge learned from pre-trained data and meanwhile focuses on current specific tasks. 

In this paper, we focus on tasks including E-commerce title summarization and review summarization. 
In generating results, the input text followed by special token is fed into the language model, then the output is generated token by token with certain stopping criteria. We notice that, generating methods matter a lot in text generation problems. We also do comparison experiments to demonstrate it. 
In real-world applications, the flexibility in controlling the length of generating outputs is needed in some situations. We propose a novel method in controlling the length of outputs to meet different length requirements by using the same language model but different generating methods.

\section{Experiments}

In this section, we compare our approach with several baselines. The symbol $^*$ in the following tables indicates that results come from their paper.

\begin{table}[htbp]
\renewcommand\arraystretch{1.25}
\centering
\caption{Comparison of our models and related work on the LCSTS text summarization task.}
\footnotesize
\begin{tabular}{c|c|c|c}
\hline
Model & Rouge-1 & Rouge-2& Rouge-L \\ \hline

Transformer$^*$\cite{duan2019contrastive} &41.93 & 28.28 & 38.32 \\
\hline
GRET$^*$\cite{weng2020gret} & 44.77 & 30.96 & 41.21 \\
\hline
DSGPT & 40.21 & 27.24 & 39.28 \\
\hline
DSGPT\_{large} & 43.80 & 31.62 & 42.92 \\
\hline
\end{tabular}
\end{table}

\begin{table}[htbp]
\renewcommand\arraystretch{1.25}
\centering
\caption{Results on Taobao.com Product Title Summarization dataset.}
\footnotesize
\begin{tabular}{c|c|c|c}
\hline
Model & Rouge-1 & Rouge-2& Rouge-L \\ \hline
MS-Pointer\_{bi}$^*$ & 75.69 &  60.29  & 75.45 \\
\hline
Transformer & 75.86 & 64.36 & 75.93 \\
\hline
DSGPT\_{w/o\ pre-training} & 75.49 & 63.96 & 75.69 \\
\hline
DSGPT & 76.09 & 64.77 & 76.33 \\
\hline
DSGPT\_{large} & 77.22 & 66.22 & 77.46 \\
\hline
\end{tabular}
\end{table}

\begin{table}[htbp]
\renewcommand\arraystretch{1.25}
\centering
\caption{Results on in-house data of review summarization.}
\footnotesize
\begin{tabular}{c|c|c|c}
\hline
Model & Rouge-1 & Rouge-2& Rouge-L \\ \hline
Transformer & 75.38 & 70.66 &	74.50 \\
\hline
DSGPT\_{w/o pre-training} & 74.76 & 70.14 & 74.34 \\
\hline
DSGPT & 77.11 & 72.71 & 76.73 \\
\hline
\end{tabular}
\end{table}

\begin{table}[htbp]
\renewcommand\arraystretch{1.25}
\centering
\caption{Effects of generating methods on in-house data of review summarization.}
\footnotesize
\begin{tabular}{c|c|c|c}
\hline
Generating Method & Rouge-1 & Rouge-2& Rouge-L \\ \hline
Beam Search with truncation & 77.11 & 72.71 & 76.73 \\
\hline
Greedy Search with truncation & 77.61 &	73.29 &	77.23 \\
\hline
Greedy Search with automatic length control & 79.90 &	76.16 &	79.59 \\
\hline
\end{tabular}
\end{table}

\begin{table*}[htbp]
\begin{CJK*}{UTF8}{gbsn}
\renewcommand\arraystretch{1.25}
\centering
\caption{Examples of generated short titles}
\scriptsize
\begin{tabular}{cccc}
\hline
Original Title & \multirows{美国 曼哈顿 Manhattan Portage 邮差包 单肩包挎包 正品现货 \\ US Manhattan Manhattan Portage Messenger Bag \\ Shoulder Bag Satchel Factory Authentic In Stock} & \multirows{Sony索尼PSP3000 全新原装主机PSP \\ 掌上游戏机破解 掌机FC GBA 街机 \\ Sony Sony PSP3000 brand new original host PSP \\ handheld game console hacked handheld FC GBA arcade} & \multirows{ESPRIT 女童直筒款牛仔长裤-WK2A \\ ESPRIT girls straight denim long pants-WK2A}  \\
\hline
Ground Truth & \multirows{Manhattan Portage 邮差包 \\ Manhattan Portage Messenger Bag} & \multirows{索尼PSP3000掌上游戏机 \\ Sony PSP3000 handheld game console} & \multirows{ESPRIT直筒牛仔裤 \\ ESPRIT straight denim pants} \\
\hline
Transformer & \multirows{曼哈顿单肩包 \\ Manhattan Shoulder Bag} & \multirows{索尼PSP3000300030003000机 \\ Sony PSP3000300030003000 machine}  & \multirows{ESPRIT 牛仔长裤 \\ ESPRIT denim long pants} \\
\hline
DSGPT\_{w/o pre-training} & \multirows{Manhattan 单肩包 \\ Manhattan Shoulder Bag } & \multirows{索尼 掌上游戏机 \\Sony handheld game console} & \multirows{ESPRIT直筒牛仔长裤 \\ ESPRIT straight denim long pants } \\
\hline
DSGPT & \multirows{Manhattan Portage 邮差包 \\ Manhattan Portage Messenger Bag} & \multirows{索尼PSP3000掌上游戏机 \\ Sony PSP3000 handheld game console} & \multirows{ESPRIT女童牛仔长裤 \\ ESPRIT girls denim long pants} \\

\hline
\end{tabular}
\end{CJK*}
\end{table*}

\subsection{Datasets}

We conduct experiments on both public and in-house datasets. We use LCSTS\cite{hu2016lcsts} as the dataset in pre-training process. For title summarization task, we use Taobao.com Production Title Summarization dataset\cite{sun2018multisource}. For review summarization task, we use a human-labeled in-house dataset from JD.com. 

\textbf{LCSTS Dataset.} Chinese short text summarization dataset(LCSTS) is collected from the Chinese microblogging website SinaWeibo. 
The split of dataset follows the original paper, with 2.4M source summary pairs as training set and 725 pairs with high annotation score as testing set. 

\textbf{Production Title Summarization Dataset.} The Production Title Summarization Dataset is collected by Alibaba. The data is crawled from the human generated pairs from a product recommendation channel of Taobao.com, and manually written by professional editors. The dataset has a regular corpus and a big corpus. We follow the data split of original paper, and use the regular corpus only as in original work, while ignoring the background knowledge part of the data. 

\textbf{JD.com Review Summarization Data.} The review summarization dataset is collected from our in-house data. The reviews come from our real customer review. The summarization of reviews are obtained by rule-based methods and human-label. To meet different requirements, the human-labeled data has both long and short versions for validation set, while only long version is provided in training set.

\subsection{Results on Chinese Short Text Summarization}

We pre-train our language models on LCSTS dataset. We train models with different sizes, DSGPT is equivalent to the
original GPT with 12 layers, and DSGPT\_{large} is equivalent to the BERT\_{large} with 24 layers, which are much smaller than the GPT-2 model with 48 layers. 
To further evaluate the performance of the pre-trained models, we generate results using beam search algorithm with beam size of 5 for the testing set. 
Table 1 presents Rouge-1, Rouge-2 and Rouge-L scores on LCSTS dataset for our model with other published methods. 
DSGPT pre-trained model achieves parity with SOTA results on the LCSTS dataset, while the other methods are not well suitable for downstream tasks since their model structures are not as flexible as ours. 
Results also show that performance of decoder-only transformer text generator can be easily improved by increasing the model size. 

\subsection{Results on Product Title Summarization}

We evaluate our method on product title summarization dataset.  We obtain our model for title summarization by fine-tuning the model obtained from Section 4.2 on product title summarization dataset. The results are generated with beam search algorithm with beam size of 5. Table 2 shows Rouge-1, Rouge-2 and Rouge-L scores on different methods. The fine-tuned DSGPT exceeds performance of machine translation model and the original work\cite{sun2018multisource}. 
In contrast with the MS-Pointer method, our model abandons additional background information but still achieves better results, which means lots of additional work in data collection and filtering can be saved in real-world applications. 
To demonstrate effects of pre-trained model, we also train a decoder transformer without initializing the weights from pre-trained model and compare the performance. 
Results show domain-specific pre-training helps increase the performance significantly.  
We also try fine-tuning on DSGPT\_{large}, results show that performance can be easily improved via scaling up the model.

\subsection{Results on Review Summarization}

\subsubsection{Comparison of different methods}

We evaluate our method on in-house review summarization dataset.  We obtain our model for review summarization by fine-tuning the model obtained from Section 4.2 on review summarization dataset. Table 3 shows Rouge-1, Rouge-2 and Rouge-L scores on different methods. The fine-tuned DSGPT exceeds performance of machine translation model and the DSGPT model without pre-training. The results are generated with beam search algorithm with beam size of 5. 

\subsubsection{Effects of controlling method in text generation}

In addition, we experiment on generating methods. The purpose is to reuse the same language model to generate summary with different length requirements. Instead of simple truncation, we introduce an automatic method in generating process. We multiply an amplification factor on the end token while generating the results token by token, that is to increase the probability of the end token in a natural way. Table 4  compares the Rouge-1, Rouge-2 and Rouge-L scores of generated results utilizing automatic method and truncate method, and results with automatic length-shorten method show improvement compared with regular truncate method.

\subsection{Case Studies}

\begin{CJK*}{UTF8}{gbsn}
To better illustrate what each model learns, we compare some examples of generated titles from each model. As show in Table 5, the fine-tuned DSGPT model performs better than the other models. 
As discussed in \cite{sun2018multisource},
summarization models typically require additional background knowledge to extract brand names and commodity names accurately from the title.
In contrast, DSGPT model achieves high summarization accuracy for brand and commodity name even when the background information is absent.
In the first column of Table 5, DSGPT model consistently includes the brand name (Manhattan Portage) and commodity name (邮差包/Messenger Bag) in the final summary, thanks to the background knolwedge learned during pre-training. No additional data is needed in fine-tuning and prediction processes. 

More generally, background knowledge from pretraining can help making the right trade-off between attributes. In the last column of Table 5, the ground truth fails to highlight the important attribute of the product (女童/girls), despite the common knowledge that (女童/girls) should be more important than the attribute (直筒/straight). Our fine-tuned DSGPT model correctly favors the former attribute which matches human reasoning more closely. 

In summary, the domain specific pre-training step is very helpful in summarizing key information of product title, while at the same time avoids the additional work on background information collection for future product titles.

\end{CJK*}

\section{Conclusion}

The proposed domain-specific pre-training combining with the decoder-only module helps improve the quality of text generation significantly. 
The decoder-only text generator is well-suited for fine-tuning tasks in text generation due to its easy-to-reuse structure and easy-to-scale-up property.
The generalized knowledge learned from pre-training helps avoid prohibitive cost of data-labeling work for additional knowledge in downstream tasks, providing a practical solution to real-world problems. This idea can be widely applied in a various of fields and languages especially when there is a shortage of data or public pre-trained models.

\bibliographystyle{ACM-Reference-Format}
\bibliography{sigir_dsgpt}

\end{document}